# Understanding the Nature of Generative AI as Threshold Logic in High-Dimensional Space

Ilya Levin

School of Computer Science, Holon Institute of Technology, Holon, Israel

levini@hit.ac.il

**Abstract**

This paper deals with the role of threshold logic in understanding the nature of generative artificial intelligence. Threshold functions, originally studied in the 1960s as elements of digital circuit synthesis, provide the most structurally transparent model of what a neural network computes: a weighted sum of inputs compared to a threshold, geometrically realized as a hyperplane partitioning a space. The paper shows that this elementary operation undergoes a qualitative phase transition when the dimensionality of the input space increases. In low dimensions, the perceptron functions as a determinate logical statement: it either separates two classes or it does not, and this is decided exactly by linear programming. In high dimensions, by contrast, a single hyperplane can separate almost any configuration of points (Cover, 1965), the space becomes saturated with potential classifiers, and the perceptron transforms from a logical device into a navigational one — an indexical indicator in the sense of Peirce's semiotics. The limitations of the perceptron identified by Minsky and Papert (1969) were perceived as a problem easily resolved by increasing the number of layers in the network, and the field moved accordingly toward multilayer architectures. However, the alternative path — increasing dimensionality while retaining a single threshold element — yields equally profound consequences for understanding neural computation. The role of multilayer architectures in high-dimensional spaces is also examined: depth serves as a mechanism for the sequential deformation of data manifolds through iterated threshold-function folds, preparing them for the linear separability that high-dimensional geometry already affords. The resulting triadic account — threshold function as ontological unit, dimensionality as enabling condition, and depth as preparatory mechanism — provides a unified explanation of generative AI, grounded in established mathematics and rooted in the historical tradition of threshold logic.

**Keywords:** threshold logic, high-dimensional geometry, perceptron, phase transition, generative AI, manifold untangling, indexicality, linear separability

## 1. Introduction

The arrival of generative artificial intelligence has posed a challenge that is not merely technical but epistemological. Systems built on neural network architectures — large language models, diffusion-based image generators, multimodal transformers — produce outputs of a quality and coherence that was, until recently, considered the exclusive province of human cognition. Yet when one asks why these systems work — not how they are engineered, but what makes their operation possible in principle — the available answers remain unsatisfying. The dominant explanations are either too technical (descriptions of architectures, training procedures, and optimization algorithms that account for the "how" but not the "why") or too vague (invocations of "emergent properties" or "statistical patterns" that gesture at an answer without providing one).

This paper originates in a sustained attempt to find a more adequate account — one that is both mathematically grounded and epistemologically transparent. The search has passed through two stages, each of which produced a partial answer, and the present work represents their synthesis.

The first stage began with the observation that the key to understanding generative AI lies not in the architecture of particular networks but in the geometry of the spaces in which they operate. High-dimensional spaces possess a constellation of properties — concentration of measure, quasi-orthogonality of random vectors, exponential capacity for linear separation, regularity of data manifolds — that collectively create an environment qualitatively different from the low-dimensional spaces of everyday intuition. These properties were analyzed in (Levin, 2026), which argued that the four geometric features of high-dimensional space constitute the foundation upon which generative computation rests. This geometric account provided explanatory power: it showed where generative AI operates and what properties of that space make its operation possible. But it left the ontology of neural networks — the question of what the network is, structurally — largely in the background.

The second stage addressed this gap by returning to a nearly forgotten tradition: threshold logic. Developed in the 1960s by Varshavsky (1962), Nechiporuk (1964), Pospelov (1974), and Muroga (1971), threshold logic established a formal identity between the artificial neuron and the threshold logic gate — both compute a weighted sum of inputs compared to a threshold, geometrically realized as a hyperplane partitioning a Boolean hypercube. We proceed from the view that this identity is not merely a formal convenience but reflects the ontological foundation of neural networks: the threshold function is what a neural network is, at its most elementary level. This tradition, oriented toward hardware synthesis rather than machine learning, preserved a structurally transparent, geometrically interpretable understanding of neural computation. In (Levin & Talis, 2026) this tradition was recovered and demonstrated its educational and epistemological value. However, that work dealt with threshold functions in low-dimensional spaces and with non-learning (fixed-weight) networks — a deliberate and acknowledged limitation.

Two questions remained open after these two stages. First: if the geometric properties of high-dimensional space are the foundation of generative AI, and if threshold functions are the ontological building blocks of neural networks, then how do threshold functions behave in high-dimensional space? Do they retain their logical character, or does something qualitatively different emerge? Second: how can a pretrained network — a system with frozen weights, structurally identical to a threshold circuit — respond so differently to different inputs, exhibiting what can only be described as indexical behavior: sensitivity to "here, now, and with me"?

The present paper provides answers to both questions, and shows that the answers are connected. The threshold function, when placed in a high-dimensional space, undergoes a phase transition: from a logical device (a symbol in the semiotic sense of Peirce) to a navigational indicator (an index). In low dimensions, a perceptron either finds a separating hyperplane or fails — a binary, all-or-nothing affair governed by linear programming. In high dimensions, the same perceptron can separate almost any configuration, and the space is saturated with potential classifiers. The character of the computation changes from deduction to navigation: not "is this separable?" but "which direction to go?"

This phase transition, governed by a single parameter — the dimensionality of the space — constitutes the ontological bridge between symbolic AI and generative AI. It is not a metaphor or a philosophical analogy; it is a mathematical fact, grounded in Cover's theorem (1965), the concentration of measure phenomenon (Ledoux, 2001), and the exponential capacity of high-dimensional spaces (Gorban & Tyukin, 2018).

A historical observation sharpens the significance of this result. When Minsky and Papert (1969) demonstrated the limitations of the perceptron — most famously, its inability to compute XOR — the difficulty was perceived as easily solvable by adding layers to the network. The field moved accordingly, and the subsequent development of multilayer perceptrons, backpropagation (Rumelhart, Hinton, & Williams, 1986), and deep learning followed this path. But there was another path, and it was not taken: instead of making the network deeper, one could have made the space wider. XOR is impossible in two dimensions; in two thousand dimensions, it is trivial. Nobody, in the early 1970s or today, proposed solving XOR by increasing the number of coordinates rather than the number of layers. Yet this is, in essence, what modern neural networks do: their embedding layers project data into high-dimensional spaces where perceptron freedom — the near-universal availability of linear separation — renders the original difficulty moot.

This does not mean that depth is unnecessary. Multilayer architectures are essential, but their function is different from what the classical account suggests. Depth does not primarily introduce nonlinear decision boundaries; it progressively deforms the manifolds on which real data lies — through iterated applications of threshold-function folds — until these manifolds are simple enough for a single hyperplane to separate. Depth does not make the classifier more complex; it makes the data more simple. The relationship between dimensionality and depth — two independent parameters, each offering a distinct path from the perceptron's limitations to the capabilities of modern AI — is illustrated schematically in Figure 1 (see Section 3).

The paper thus proposes a triadic account of generative AI: the threshold function as the ontological unit; dimensionality as the enabling condition (providing perceptron freedom); and depth as the preparatory mechanism (deforming manifolds to realize that freedom). Together, these three elements constitute a unified explanation — grounded in established mathematics, rooted in the historical tradition of threshold logic, and connecting the symbolic and generative paradigms through a single continuous parameter. In this way, the paper connects two prior lines of work: the geometric epistemology of generative AI (Levin, 2026) and conceptualizing neural networks as logical structures of threshold logic (Levin & Talis 2026), positioning the dimensional phase transition of threshold functions as the ontological bridge between symbolic and generative artificial intelligence.

The paper is organized as follows. Section 2 examines the perceptron as a logical statement in low-dimensional space, tracing the development from McCulloch–Pitts through Rosenblatt and Minsky–Papert to the threshold logic tradition. Section 3 analyzes the perceptron in high-dimensional space, presenting Cover's theorem, concentration of measure, and the concept of perceptron freedom. Section 4 addresses the role of depth as manifold transformation, explaining how layers of threshold functions progressively simplify data geometry. Section 5 develops the semiotic interpretation — the transition from symbol to index — and its consequences for understanding pretrained generative systems. Section 6 discusses implications for explainability, hallucination, and open questions.

## 2. The Perceptron as a Logical Statement

The argument of this paper begins with the simplest object in neural computation: a single threshold function. Before examining what happens to this object in high-dimensional space (Section 3) or in multilayer compositions (Section 4), we must establish precisely what it is and what it does in the low-dimensional regime where it was originally studied. This section traces the historical development of the perceptron as a logical device, from its origins through its celebrated limitations to the threshold logic tradition that preserved its structural-geometric understanding.

### 2.1 Origins: McCulloch–Pitts and the Logical Neuron

The conceptual identification of neurons with logical elements predates the modern era of artificial intelligence. In their landmark 1943 paper, McCulloch and Pitts proposed that the activity of a neuron could be modeled as a logical proposition: a neuron either fires or does not, and its firing is determined by the weighted sum of its inputs exceeding a threshold (McCulloch & Pitts, 1943). This model, despite its severe biological simplifications, established a correspondence between neural activity and mathematical logic that proved extraordinarily fertile.

The McCulloch–Pitts neuron computes a function of the form:

$$y = 1 \text{ if } \Sigma w_i x_i \geq \theta, \text{ and } y = 0 \text{ otherwise}$$

where $x_i \in \{0, 1\}$ are binary inputs, $w_i$ are integer weights, and $\theta$ is a threshold value. As demonstrated systematically in the threshold logic literature (Muroga, 1971; Levin & Talis, 2026), this simple device can realize a substantial class of Boolean functions — all functions of one variable, fourteen of the sixteen functions of two variables (the exceptions being XOR and its complement), and all monotone increasing functions of up to three variables.

The crucial insight — one that connects neural computation to classical logic — is that a threshold function partitions the vertices of the Boolean hypercube $\{0, 1\}^n$ into two classes by means of a hyperplane. A Boolean function is a threshold function if and only if it is linearly separable: if there exists a hyperplane in $\mathbb{R}^n$ that correctly separates the vertices where the function evaluates to 1 from those where it evaluates to 0. The geometric representation is direct and vivid: in two dimensions, the hyperplane is a line dividing the four vertices of a square; in three dimensions, it is a plane cutting through the eight vertices of a cube (Levin & Talis, 2026, Figures 1–6).

### 2.2 Rosenblatt and the Perceptron: From Logic to Learning

Frank Rosenblatt's perceptron (1958, 1962) transformed the McCulloch–Pitts model from a static logical device into a learning system. While the underlying computation remained the same — a weighted sum compared to a threshold — Rosenblatt introduced a procedure for adjusting the weights on the basis of training examples. The perceptron convergence theorem established that, for any linearly separable training set, this procedure would converge to a correct set of weights in a finite number of steps. The perceptron thus demonstrated that a threshold element could not only compute logical functions but could discover the appropriate weights automatically — a capability that would, decades later, be scaled to the training of networks with billions of parameters.

**2.3 Minsky–Papert and the Limits of Linear Separability**

The optimism surrounding the perceptron was abruptly curtailed by Minsky and Papert's *Perceptrons* (1969), one of the most consequential publications in the history of artificial intelligence. Their central result was a systematic demonstration of the functions that a single perceptron cannot compute. The most celebrated example is XOR: the function $f(x_0, x_1) = x_0 \oplus x_1$, which evaluates to 1 when exactly one of its inputs is 1 and to 0 otherwise.

The impossibility of XOR for a single perceptron is a direct consequence of linear separability. The four vertices of the unit square in $\mathbb{R}^2$ where XOR evaluates to 1 — namely (0,1) and (1,0) — and where it evaluates to 0 — namely (0,0) and (1,1) — cannot be separated by any line. The "positive" vertices lie on one diagonal of the square, and the "negative" vertices on the other; no line can separate the two diagonals.

Minsky and Papert extended this observation far beyond XOR, demonstrating fundamental limitations in the perceptron's ability to compute predicates involving connectivity, parity, and other topological or global properties. Their analysis was mathematically impeccable and their conclusion was clear: single-layer networks are fundamentally limited, and these limitations are consequences of the geometry of linear separation.

**2.4 The Historical Choice: Depth over Dimensionality**

The response of the research community to Minsky and Papert was to pursue architectural complexity. If a single perceptron cannot compute XOR, then use two perceptrons in a hidden layer and one in an output layer. This reasoning led, over the subsequent decades, to the development of multilayer perceptrons (Rumelhart, Hinton, & Williams, 1986), convolutional networks (LeCun et al., 1998), and eventually the deep architectures that characterize contemporary AI.

The classical understanding of what multilayer architectures accomplish is well known (Haykin, 2009): a single-layer network can implement half-planes bounded by hyperplanes; a two-layer network can implement convex open or closed regions; a three-layer network can implement arbitrary regions, limited only by the number of nodes. Each additional layer adds topological complexity to the decision boundary.

This understanding is correct, but it conceals an implicit assumption: that the dimensionality of the input space is fixed. XOR is hard in two dimensions. The entire multi-layer perceptron tradition operates within a framework where the number of input features is a given, and the architecture must be chosen to handle whatever complexity the data presents in that fixed space.

What was not pursued is the alternative: change the dimensionality of the space. This alternative was, in a sense, implicit in the work of the threshold logic tradition — Varshavsky (1962), Pospelov (1974), and their collaborators had developed a comprehensive theory of threshold circuits in spaces of various dimensions — but this work was oriented toward hardware synthesis and logical design, not toward the learning-theoretic questions that Minsky and Papert had foregrounded. The result was a historical bifurcation: one path led to deeper networks, the other to a forgotten corner of digital logic design. The present paper argues that the second path leads to insights essential for understanding why modern generative AI works.

## 2.5 The Threshold Logic Tradition

The tradition of threshold logic deserves more extended discussion than it typically receives in contemporary AI literature, both for historical justice and because its concepts are central to our argument.

The theoretical foundations were laid in the early 1960s, in parallel with Rosenblatt's work on perceptrons but within the framework of logical synthesis rather than machine learning. Varshavsky (1962) published an essential study on logical networks constructed from threshold elements. Nechiporuk (1964) provided formal complexity estimates for threshold circuits. Pospelov (1974) consolidated and advanced this body of work in his monograph on logical methods for circuit analysis and synthesis. In the English-language literature, Muroga (1971) provided a comprehensive formalization of threshold logic, including analytical methods for determining whether a given Boolean function is a threshold function and algorithms for finding the associated weights and threshold.

The key insight of this tradition is the formal identity between the artificial neuron and the threshold logic gate. Both compute the same function: a weighted sum of binary inputs compared to a threshold. But the two traditions drew different consequences from this identity. The neural network tradition, following Rosenblatt, emphasized learning: the ability to find weights by exposure to examples. The threshold logic tradition, following Varshavsky and Muroga, emphasized synthesis: the ability to construct circuits with desired logical properties. The former led to gradient descent, backpropagation, and deep learning. The latter led to CMOS implementations, β-driven threshold elements, and hardware-efficient neural network design (Varshavsky, Marakhovsky, & Levin, 2003).

As have argued in (Levin & Talis, 2026), the threshold logic tradition preserved a mode of understanding neural computation that is structural and geometric, as opposed to the procedural and algorithmic understanding that dominates contemporary machine learning. The geometric transparency of threshold functions — the fact that they can be visualized as hyperplanes cutting through hypercubes — provides an epistemic accessibility that is lost in the opacity of deep architectures. It is precisely this structural-geometric mode that the present paper seeks to extend into high-dimensional spaces.

A point of particular importance for our argument concerns the mathematical character of threshold function synthesis. The problem of finding weights and a threshold for a given classification task is a problem of linear programming: find a weight vector and a scalar threshold such that the weighted sum exceeds the threshold for all positive examples and falls below it for all negative examples. This is a system of linear inequalities, and its feasibility can be determined exactly by the methods of linear programming (Dantzig, 1963). The significance of this characterization is that it anchors the perceptron in a domain of exact, provably correct computation. There is no approximation, no heuristic, no gradient descent into local minima. Either the system of inequalities has a solution and the data is linearly separable, or it does not and no single perceptron can solve the problem. The perceptron, in this regime, operates as a device of mathematical logic: its judgments are binary, its correctness is provable, and its limitations are absolute. This mathematical exactness — the fact that threshold function synthesis is a solved problem in the precise sense of linear programming — is what gives our subsequent argument about the phase transition its force: we are not tracking the behavior of an approximate or heuristic device, but of a mathematically exact one.

An important technical note: the equivalence between neural networks and threshold circuits holds rigorously for non-learning (fixed-weight) networks. For learning networks, the dynamics of weight adjustment introduce considerations beyond the combinatorial structure of threshold functions. Our argument in Sections 3 and 5, which concerns the behavior of pretrained (and therefore fixed-weight) networks, relies on this non-learning equivalence and is therefore on firm ground.

Summarizing, in low-dimensional space, the perceptron is a logical device. It computes a threshold function — a Boolean function realizable by a hyperplane — and the problem of finding the appropriate hyperplane is a problem of linear programming. Its capabilities and limitations are precisely characterized. The perceptron in low dimensions operates in a world of binary certainties — correct or incorrect, separable or inseparable.

## 3. The Perceptron in High-Dimensional Space

We now ask: what happens to this precisely characterized logical instrument when the space in which it operates grows from two or three dimensions to thousands or millions? The answer is not that the perceptron becomes "better" at the same task, but that the nature of its task changes.

### 3.1 Cover's Theorem and the Geometry of Linear Separability

The transformation is captured with mathematical precision by Cover's theorem on the capacity of linear classifiers (Cover, 1965).

Consider $N$ points in general position in $\mathbb{R}^n$. A *dichotomy* is any assignment of these points to two classes. Cover's theorem states that the number of linearly separable dichotomies — binary labelings realizable by a single hyperplane is:

$$C(N, n) = 2 \sum_{k=0}^{n-1} \binom{N-1}{k}$$

The critical consequence: when $N \leq 2n$, the fraction of realizable dichotomies approaches 1 as the dimension grows. A single hyperplane in $\mathbb{R}^n$ can correctly classify almost any configuration of up to $2n$ points into any desired binary partition. For $N$ much larger than $2n$, this fraction drops sharply toward zero — a phase transition in the classical statistical-mechanical sense.

To appreciate the magnitude: in $\mathbb{R}^2$, a single line can correctly separate at most 4 points in general position, and the XOR arrangement is already beyond its reach. In $\mathbb{R}^{1000}$, a single hyperplane can correctly separate virtually any configuration of up to 2000 points. In $\mathbb{R}^{10000}$, the capacity extends to 20000 points. The same mathematical object — a linear inequality — that fails at XOR in two dimensions becomes a nearly universal classifier in high dimensions.

It is worth pausing to explain why additional dimensions resolve the XOR problem, since this point is central to our argument. In $\mathbb{R}^2$, the four points of XOR — (0,0), (0,1), (1,0), (1,1) — lie flat on a plane. The two "positive" points (0,1) and (1,0) sit on one diagonal; the two "negative" points (0,0) and (1,1) sit on the other. No line can separate one diagonal from the other — this is the geometric content of Minsky and Papert's impossibility result. But now suppose we add a third coordinate — say, $x_2 = x_0 \cdot x_1$ (the product of the two original inputs). The four points lift into $\mathbb{R}^3$: (0,0) becomes (0,0,0), (0,1) becomes (0,1,0), (1,0) becomes (1,0,0), and (1,1) becomes (1,1,1). The "negative" point (1,1,1) has moved upward along the new axis, away from the other three, and a plane can now easily separate (0,1,0) and (1,0,0) from (0,0,0) and (1,1,1). The additional dimension has provided the geometric room that was absent in $\mathbb{R}^2$.

A natural objection arises: for XOR, a single additional dimension suffices — three dimensions, not three thousand. Why, then, do we speak of high-dimensional spaces? The answer is that XOR is merely the simplest precedent — the first case in which a Boolean function of two variables is unrealizable by a single perceptron. In reality, such functions are numerous, and their number grows rapidly with the number of variables. At two variables, only 2 of the 16 possible Boolean functions are not threshold functions (XOR and its complement). At four variables, functions like $x_0 x_1 + x_2 x_3$ are already beyond the perceptron's reach (Levin & Talis, 2026). As the number of variables increases, the fraction of Boolean functions that are threshold functions shrinks dramatically — the vast majority of Boolean functions are not linearly separable (Muroga, 1971). This is the real scope of the problem that Minsky and Papert identified: not merely XOR, but an exponentially growing class of functions that resist single-hyperplane classification. And it is this full scope that requires genuinely high-dimensional spaces. Cover's theorem provides the resolution: by increasing dimensionality sufficiently — not by one or two dimensions, but by orders of magnitude — almost any configuration of points becomes linearly separable, regardless of how entangled it was in the original space.

This is the first component of what we term perceptron freedom: in high-dimensional space, the perceptron is freed from the constraints that define its behavior in low dimensions. The question is no longer whether a separating hyperplane exists, but which among exponentially many possible hyperplanes to choose.

### 3.2 Concentration of Measure and the Structure of High-Dimensional Space

Cover's theorem draws its power from a constellation of geometric phenomena that characterize high-dimensional spaces — phenomena discovered independently in probability theory, statistical mechanics, and convex geometry, unified under the rubric of "concentration of measure" (Ledoux, 2001; Vershynin, 2018).

Concentration of measure. On the unit sphere $S^{n-1}$ in $\mathbb{R}^n$, the uniform distribution becomes increasingly concentrated around any equator as n grows. The observable diameter of the sphere — the range of values that a Lipschitz function takes on most of the sphere — shrinks as $1/\sqrt{n}$, even though the metric diameter remains constant (Lévy, 1951; Milman, 1971). For a perceptron, this means that the weighted sum $\Sigma w_i x_i$ — being a Lipschitz function — is itself concentrated around its mean on high-dimensional data. Most data points produce similar values, and the hyperplane encounters most of the data near its surface.

Quasi-orthogonality. In $\mathbb{R}^n$, two randomly chosen unit vectors have an inner product whose expected square is 1/n (Vershynin, 2018). As n grows, random vectors become nearly orthogonal with high probability: in $\mathbb{R}^{10000}$, the standard deviation of angles between random pairs is approximately 0.6°. For a system of perceptrons, this means that the normal vectors of different hyperplanes are almost certainly nearly independent: each threshold function points in a direction almost orthogonal to every other. A large collection of perceptrons can encode exponentially many distinct classifications without mutual interference.

Exponential capacity. The number of nearly orthogonal vectors that can be packed into $\mathbb{R}^n$ grows exponentially with n: for any $\varepsilon > 0$, there exist exp(cn) unit vectors with all pairwise inner products bounded by $\varepsilon$ (Gorban & Tyukin, 2018; Gorban, Makarov, & Tyukin, 2020). This is the "blessing of dimensionality" (Donoho, 2000): high-dimensional spaces possess structural regularities that make certain computational tasks easier than their low-dimensional counterparts. The space is not merely large; it is dense with potential classifiers.

The foregoing properties allow us to articulate the phase transition explicitly.

In low-dimensional space (n small), the perceptron operates in a regime of scarcity. Separating hyperplanes are rare; most configurations are not linearly separable; XOR is typical. Finding a separating hyperplane, when one exists, is a constrained optimization problem with few solutions. The perceptron is a precise logical instrument.

In high-dimensional space (n large), the perceptron operates in a regime of abundance. Almost any configuration is linearly separable; separating hyperplanes are ubiquitous; the choice is vastly underdetermined by the data. The perceptron is no longer a logical instrument but a navigational one: it indicates a direction within a space of exponentially many possible classifications.

The transition is sharp — a genuine phase transition in the sense that the fraction of separable dichotomies changes from near-zero to near-one over a narrow range of the ratio N/n. This sharpness is not a mathematical curiosity; it is the quantitative signature of a qualitative change in the nature of the computation that the perceptron performs.

The analogy to physical phase transitions is more than metaphorical. Concentration of measure has its origins in statistical mechanics, where it governs the equivalence of ensembles in the thermodynamic limit (Gibbs, 1902). The same mathematical structure that explains why macroscopic quantities are well-defined despite microscopic chaos also explains why a single hyperplane can reliably classify in high dimensions despite the complexity of the data.

We are now in a position to state precisely the historical observation that motivates our reinterpretation. When Minsky and Papert demonstrated the limitations of the perceptron in 1969, the XOR problem was perceived as requiring a fundamental architectural response — the addition of hidden layers. Yet the alternative was always available: instead of making the network deeper, make the space wider.

This is, in fact, precisely what modern neural networks do — though the connection to threshold logic was not articulated. The embedding layers of a transformer, the convolutional feature maps of a CNN, the random projections of kernel methods — all are mechanisms for increasing the effective dimensionality of the representation, placing data in a space where the perceptron's linear separation is sufficient. The solution to Minsky and Papert's challenge was always implicit in the geometry of the space itself. The field needed only to ask the right question: not "how can we build more complex classifiers?" but "in how many dimensions does a simple classifier become sufficient?"

The two paths from the perceptron — increasing dimensionality versus increasing depth — and their convergence in generative AI are summarized in Figure 1.

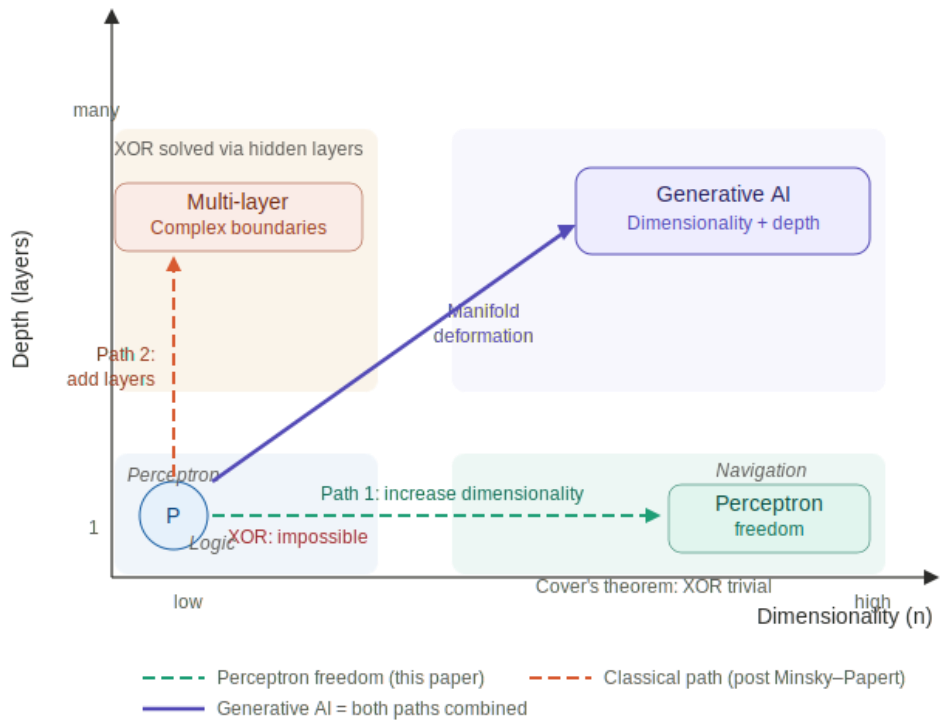

**Figure 1.** Two paths from the perceptron.

In Fig. 1, the horizontal axis represents the dimensionality of the input space; the vertical axis represents the depth (number of layers) of the network. The perceptron (P) sits at the origin: a single threshold function in low-dimensional space, where XOR is impossible. The green dashed path (increasing dimensionality at depth 1) leads to perceptron freedom — the regime where Cover's theorem guarantees near-universal linear separability. This is the path not taken after Minsky and Papert. The orange dashed path (increasing depth at low dimensionality) is the classical path: multilayer perceptrons with increasingly complex decision boundaries. Generative AI (purple diagonal) requires both: high dimensionality provides the space of possibilities, while depth provides the mechanism for manifold deformation that realizes those possibilities. The transition from the lower-left (logic) to the upper-right (navigation) is the phase transition from symbol to index that this paper describes.

## 4. Depth as Manifold Transformation

If high dimensionality alone provides perceptron freedom, then what is the function of the dozens or hundreds of layers in a modern neural network? The analysis of Section 3 sharpens this question: if a single hyperplane in $\mathbb{R}^{10000}$ can separate almost anything, why does a modern transformer need 96 layers? This section provides the answer: the distinction lies between points in general position and data on manifolds.

**4.1 Points versus Manifolds**

Cover's theorem applies to points in general position — points whose configuration is unconstrained. Real-world data, however, is not in general position. Images, texts, audio signals, and other natural data lie on low-dimensional manifolds embedded in high-dimensional ambient spaces — a principle known as the manifold hypothesis (Bengio, Courville, & Vincent, 2013; Fefferman, Mitter, & Narayanan, 2016).

The manifold hypothesis has been empirically validated across multiple domains. The space of all possible 256×256 pixel images have dimension 196608, but the set of natural photographs of human faces occupies a manifold of far lower intrinsic dimension — perhaps a few hundred — within this space (the degrees of freedom corresponding to pose, expression, lighting, identity, and a limited number of other factors). Similarly, the space of all possible token sequences is astronomical, but meaningful English sentences occupy a low-dimensional manifold within it.

A manifold is not a collection of isolated points; it is a continuous, curved, potentially self-intersecting surface. Two manifolds representing different classes — say, images of cats and images of dogs — may be intertwined in the ambient space in ways that make them linearly inseparable. Not because the space lacks capacity (it does not — perceptron freedom guarantees this), but because the manifolds' geometry is too complex for a single hyperplane to disentangle them.

This distinction is crucial. Cover's theorem says: "random points can almost always be separated." The manifold hypothesis says: "real data is not random points." The gap between these two statements is precisely where depth enters.

**4.2 The Elementary Operation: Folding by Threshold Functions**

To understand how layers bridge this gap, we return to the threshold function as our elementary unit of analysis — maintaining the connection to threshold logic that runs throughout this paper.

Each neuron in a layer computes a threshold function: a weighted sum followed by a nonlinear activation. In modern networks, the most common activation is ReLU (Rectified Linear Unit), defined as $f(x) = \max(0, x)$: it outputs the input unchanged if it is positive, and outputs zero otherwise (Nair & Hinton, 2010). Despite its simplicity — ReLU is merely a threshold at zero — this activation function has become the standard in deep learning due to its computational efficiency and its favorable properties for gradient-based training. What does a single ReLU neuron do to a manifold geometrically?

It folds it. The hyperplane defined by the weighted sum $\Sigma w_i x_i = \theta$ divides the space in two. The ReLU activation maps all points on one side of this hyperplane to zero (collapsing them onto the hyperplane), while leaving points on the other side unchanged. The effect on a manifold passing through both regions is a fold: the manifold is bent along the hyperplane, like a sheet of paper creased along a straight line.

This geometric picture can be made precise. Consider a smooth curve (a one-dimensional manifold) in $\mathbb{R}^2$. If the curve crosses a line (the hyperplane of a single neuron), the ReLU activation folds the portion of the curve on the negative side onto the line, while the portion on the positive side remains in place. The curve, formerly smooth, now has a kink at the crossing point. In higher dimensions, the same operation folds a manifold of any dimensionality along a hyperplane of codimension one.

A single fold accomplishes little — it introduces one crease. But a layer of n neurons performs n simultaneous folds along n different hyperplanes. The manifold is creased from n directions at once. If n is large (as it is in modern networks, where layer widths of 4096 or more are common), the cumulative effect is a substantial compression and restructuring of the manifold's geometry.

The entire process — from intertwined manifolds through successive folds to final linear separability — is illustrated in Figure 2.

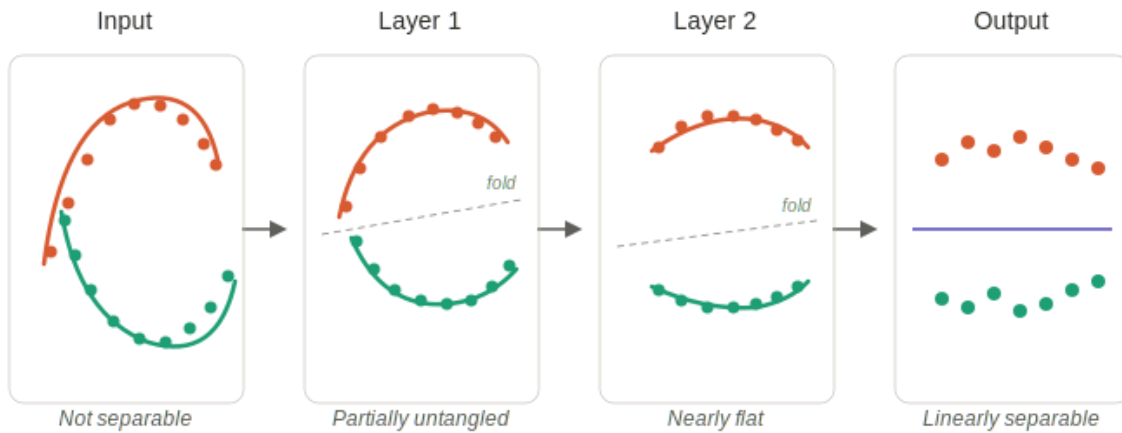

**Figure 2.** Manifold deformation through layers of threshold functions.

Fig 2 illustrates two interleaved manifolds — represented as coral and teal half-moons — corresponding to data from two classes that are not linearly separable in the input space (left panel). Each dashed line in the subsequent panels represents a single threshold function, i.e., a hyperplane along which the ReLU activation performs its fold. As the data passes through Layer 1 and Layer 2, these folds progressively simplify the manifolds' geometry: curvature decreases, the two classes are gradually pulled apart, and the intrinsic dimensionality of each manifold is reduced. By the output layer (right panel), the formerly intertwined manifolds have been deformed into compact, well-separated clusters that a single hyperplane (the purple line) can divide. This is the geometric meaning of depth: the final layer's classifier remains a single threshold function — the simplest possible operation — but the preceding layers have prepared the data so that this simple operation suffices.

The output of one layer becomes the input to the next, and the process repeats. Each layer takes the manifold as deformed by all previous layers and applies a new battery of folds. The effect is progressive simplification. Empirical studies confirm that this indeed occurs in trained deep networks. Across a range of architectures — convolutional networks, residual networks, and vision transformers — researchers have observed that manifold curvature, intrinsic dimensionality, and topological complexity all decrease from layer to layer, while classification capacity correspondingly increases (Chung, Lee, & Sompolinsky, 2018; Cohen, Chung, Lee, & Sompolinsky, 2020). By the time data reaches the final layer, the manifolds corresponding to different classes have been simplified to the point where they resemble the "points in general position" to which Cover's theorem applies. The last layer — a single threshold function, a single hyperplane — can then accomplish the separation. The entire depth of the network has worked to ensure that this final, simple operation succeeds.

A particularly illuminating theoretical contribution comes from Li (2023), who distinguished two complementary aspects of manifold untangling:

Global embedding — placing the manifold in a higher-dimensional space where entanglements can be resolved. Li noted that a knot in three dimensions can always be untied in four dimensions — a topological fact (Crowell & Fox, 2012) with direct implications for understanding why wider layers aid classification. This mechanism corresponds to what we have called perceptron freedom: the space provides enough room for disentanglement.

Local flattening — reducing curvature within each manifold through identity-preserving transformations. This mechanism corresponds to the layer-by-layer simplification described above: each layer smooths the manifold without destroying its identity structure.

Networks employ both mechanisms simultaneously. The interaction between them provides a precise vocabulary for the dual role that dimensionality and depth play in neural computation.

**4.3 Dimensionality and Depth: A Unified View**

We can now synthesize the arguments of Sections 3 and 4.

Dimensionality provides the territory — the space of possibilities. In high-dimensional space, perceptron freedom guarantees that linear separability is generically available. The four properties of high-dimensional space identified in (Levin, 2026) collectively create an environment in which threshold functions are almost universally effective.

Depth provides the route through this territory. Real data lies on complex manifolds whose geometry may obstruct the linear separability that the space nominally affords. The layers of a deep network sequentially deform these manifolds — each layer applying a battery of threshold-function folds — until the data's geometry is sufficiently simple for a final hyperplane to accomplish the separation.

Neither is sufficient alone. High dimensionality without depth provides a territory rich in potential separations but unable to access them, because the data manifolds are too entangled. Depth without high dimensionality provides a deformation mechanism but lacks the geometric room to disentangle — the regime of XOR in $\mathbb{R}^2$, where no amount of folding helps.

Generative AI requires both. And the threshold function — the single hyperplane, the elementary fold — is the common element: in high dimensions, it provides the freedom; in deep architectures, it provides the mechanism of deformation.

### 4.4 Depth Reconsidered

This unified view revises the standard account of multilayer architectures. The traditional explanation emphasizes that additional layers introduce increasingly complex nonlinear decision boundaries. Our analysis suggests a different interpretation: the nonlinearity that matters is not in the decision boundary but in the transformation of the data. Each layer applies a nonlinear map, but the purpose is to simplify the data's geometry so that a simple final boundary suffices.

Depth does not make the classifier more complex. Depth makes the data more simple. The classifier — the final hyperplane — remains a single threshold function. The data reaching it has been prepared, through layers of folding and flattening, to be amenable to exactly this type of classification.

## 5. From Threshold to Index

The preceding sections established a factual account: the threshold function behaves as a logical device in low dimensions and as a navigational instrument in high dimensions, while depth serves to prepare data for the latter regime. This section provides the interpretive framework: a semiotic analysis, grounded in Peirce's theory of signs, that articulates the precise nature of the transition and its consequences for understanding generative AI.

### 5.1 Symbol and Index: The Semiotic Framework

Charles Sanders Peirce (1931–1958) classified signs into three types based on the nature of the sign-object relation. A *symbol* signifies by convention or rule: the word "dog" denotes dogs by conventional association; logical propositions denote relations by rules of inference. Symbols are context-independent. An *index* signifies by physical or causal connection: a weathervane indicates wind direction because the wind physically moves it; a pointing finger indicates a spatial region. Crucially, an index is context-dependent: the same weathervane points differently at different times.

In low-dimensional space, the perceptron functions as a symbol. Consider the threshold function $f(x_0, x_1) = x_0 \wedge x_1$, realized by the inequality $x_0 + x_1 \geq 1.5$. This function denotes logical AND: it asserts that both inputs must be active. The hyperplane occupies a determinate position dictated by the logical content of the function. There is no ambiguity, no context-dependence. The perceptron states a proposition. The symbolic character is reflected in the mathematics: linear programming yields a determinate solution; the weights encode a specific logical relation; the geometry is fixed and interpretable.

In high-dimensional space, the character changes. A single hyperplane in $\mathbb{R}^{10000}$ can correctly classify almost any dichotomy of up to 20000 points. But which dichotomy it classifies depends entirely on the configuration of the data. The same hyperplane, in the same space, separates different classes depending on what data is presented. This is indexical behavior. The hyperplane does not assert a proposition; it indicates a direction. Its normal vector points toward one region and away from another, but the meaning of "toward" and "away" is determined by the data's position, not by the hyperplane's intrinsic properties. The hyperplane is a weathervane: it orients the system relative to whatever configuration it encounters.

### 5.2 Indexicality, Pretrained Systems, and the Ontological Bridge

This semiotic reinterpretation illuminates a puzzle about generative AI: how can a pretrained system — a network with frozen weights — respond so differently to different inputs? The answer lies in perceptron freedom. A pretrained network is a fixed structure in a high-dimensional space saturated with potential separations. The weights are fixed, but the input varies — and in high-dimensional space, every distinct input occupies a unique position relative to the network's hyperplanes. The network does not change; the data's position determines which of the network's latent distinctions are activated.

A weathervane does not change its physical structure when the wind shifts; it responds by orienting differently. A pretrained network does not change its weights when a new prompt arrives; it responds by activating a unique configuration of threshold functions — a configuration determined by the input's position in the high-dimensional space. The "here, now, and with me" quality of generative AI responses — their contextual sensitivity despite fixed weights — is a direct consequence of the geometry: in sufficiently high dimensions, no two inputs occupy the same position, and each position activates a distinct pattern of threshold functions. Indexicality is not a metaphor; it is a geometric description of how threshold functions behave in high-dimensional space.

We can now state the central thesis in its fullest form. The threshold function undergoes a phase transition as dimensionality increases: from *symbol* (logical statement) to *index* (directional indicator). This is not a replacement of one paradigm by another. Symbolic AI and generative AI are the same threshold structure operating at different points along the dimensionality axis. In low dimensions, the threshold function computes logic. In high dimensions, it navigates meaning. The ontological bridge between symbolic and generative AI is a mathematical phase transition, governed by a single parameter: the dimension of the space.

## 6. Conclusion

This paper began with an epistemological challenge: how to understand the nature of generative AI — not merely its engineering, but the structural and geometric conditions that make its operation possible. The search for an answer passed through two stages — the geometry of high-dimensional spaces and the threshold logic tradition — each of which provided a partial account. The present work has synthesized these two lines of inquiry and arrived at a unified explanation.

The central finding is that the threshold function — the simplest computational unit of neural networks — undergoes a phase transition when the dimensionality of its input space increases. In low dimensions, it functions as a logical device: a symbol that asserts a proposition, whose correctness is decided exactly by linear programming. In high dimensions, it becomes a navigational instrument: an index that points in a direction, whose meaning is determined by the context of the data. This transition, governed by a single parameter, constitutes the ontological bridge between symbolic and generative artificial intelligence.

Three elements compose the full picture. The threshold function is the ontological unit — it computes logic in low dimensions and indicates direction in high dimensions, serving as the common element across all scales of neural computation. Dimensionality is the enabling condition — high-dimensional space provides perceptron freedom, the near-universal availability of linear separation, through the mechanisms of concentration of measure, quasi-orthogonality, and exponential capacity. Depth is the preparatory mechanism — the layers of a deep network deform data manifolds through iterated threshold-function folds, progressively simplifying their geometry until the perceptron freedom of high-dimensional space can be realized.

The paper has also traced a historical path not taken. The limitations of the perceptron demonstrated by Minsky and Papert were addressed by adding layers; the alternative — increasing dimensionality — was not pursued, despite being implicit in the threshold logic tradition. Yet this is precisely what modern neural networks accomplish through their embedding layers. The irony is that the solution to the perceptron's limitations was always present in the geometry of the space itself.

Several implications follow. For explainability, the analysis suggests that the opacity of neural networks is a consequence of dimensionality rather than complexity — a single perceptron in ten thousand dimensions computes a linear function, but its behavior exceeds human spatial intuition. For hallucination, the geometric perspective reveals it as a structural feature of navigation in high-dimensional space: an indexical system always points in a direction, even when no grounded meaning corresponds to that direction. These are not engineering failures to be fixed but geometric consequences of the same structure that makes generative AI possible.

Open questions remain. The formal derivation of perceptron freedom from the four geometric properties of high-dimensional space; the bounds on minimum depth required for manifold simplification; the connections between the semiotic interpretation and philosophical debates about understanding in AI; and the practical implications for architecture design — all invite further investigation. The recent success of wide architectures and mixture-of-experts models may already be an empirical manifestation of the principle that dimensionality, rather than depth alone, is the fundamental computational resource.

The paper began with the observation that generative AI poses an epistemological challenge. It concludes with the proposal that threshold logic — a nearly forgotten tradition from the 1960s — provides the ontological key to meeting that challenge. The threshold function, the simplest element of neural computation, contains within itself the entire arc from symbolic logic to generative navigation. One needs only to increase the dimension of the space in which it operates.